\documentclass[letterpaper, 10 pt, conference]{ieeeconf}  

\IEEEoverridecommandlockouts                              

\usepackage{graphics} 
\usepackage{epsfig} 
\usepackage{makecell}

\title{\LARGE \bf
Fast and Robust Teach-and-Repeat Navigation Using MixVPR Visual Place Recognition*
}

\author{V\'aclav Truhla\v{r}\'ik$^{1, 2}$, Tom\'a\v{s} Pivo\v{n}ka$^{1}$, and Libor P\v{r}eu\v{c}il$^{1}$  ~\IEEEmembership{Member,~IEEE} 
\thanks{*This work was co-funded by the European Union under the project Robotics and advanced industrial production (reg. no. CZ.02.01.01/00/22\_008/0004590) and by the Grant Agency of the Czech Technical University in Prague, grant no. SGS25/157/OHK3/3T/13.}
\thanks{$^{1}$All authors are with Czech Institute of Informatics, Robotics and Cybernetics, Czech technical University in Prague, Jugoslávských partyzánů 1580/3, 160 00 Praha 6, Czech Republic ({\tt\small tomas.pivonka@cvut.cz})}
\thanks{$^{2}$V\'aclav Truhla\v{r}\'ik is also with the Department of Cybernetics, Faculty of Electrical Engineering,
Czech Technical University in Prague, Karlovo náměstí 13, 121 35 Praha 2, Czech Republic}%
}

\begin{document}

\maketitle
\thispagestyle{empty}
\pagestyle{empty}

\begin{abstract}
Teach-and-repeat navigation systems employing advanced visual place recognition techniques for localization exhibit key attributes for long-term mobile robot navigation, such as the ability to operate in unstructured and dynamic environments. However, existing solutions based on deep-learning techniques have high computational requirements, limiting their applicability. This work introduces a novel and efficient teach-and-repeat system built on the modern visual place recognition method MixVPR. Real-world testing demonstrated its ability to operate both indoors and outdoors, achieving robustness and navigation precision comparable to other state-of-the-art systems. In addition, its lower hardware requirements make it suitable for a wide range of robotic platforms and practical applications.
\end{abstract}

\section{INTRODUCTION}

Reliable navigation of mobile robots in unstructured and changing environments is the key area of research for long-term robot autonomy. Modern robots can be equipped with various specialized sensors; however, they are usually expensive and unsuitable for some types of applications. In addition, processing all the data requires a lot of computational power. Therefore, efficient systems working with a single forward-looking monocular camera, such as the Teach-and-Repeat (T\&R) navigation system introduced in this article,  are valuable, especially for real applications.

T\&R systems are capable of following previously traversed trajectories. They operate in two stages. At first, the desired trajectory is taught during a traversal when a~human operator or an outside algorithm controls the robot. Subsequently, in the second stage, the robot autonomously repeats the trajectory using current sensor data and information captured in the teaching phase.

There are two main approaches to T\&R navigation: position-based and appearance-based. The position-based system localizes the robot in space and guides it based on the difference to the corresponding position from a database. By contrast, appearance-based systems need only 1D localization or tracking along the taught trajectory. When localized, the shift between the current and corresponding database image is used to control the robot heading reactively.

The article focuses on the appearance-based approach with the use of advanced visual place recognition systems for localization. The work is motivated by previous systems of this type \cite{SSM-Nav, SSM-Nav2}, which proved the ability to operate under challenging lighting conditions and in low-structured environments. However, their main limitation is long computational time, which is caused by the high demands of the used visual place recognition systems.

The presented system solves this limitation by using a~more efficient one-stage Visual Place Recognition (VPR) system - MixVPR \cite{MixVPR}, reaching performance close to state-of-the-art systems \cite{SSM-VPR2}. The introduced T\&R system builds on the previously mentioned system SSM-Nav2 \cite{SSM-Nav2}. Besides the direct replacement of the original VPR method by MixVPR, a new approach to horizontal shift computation is introduced. It takes advantage of MixVPR architecture and extracts local visual features from its convolutional neural network backbone. Capabilities of the new T\&R system were tested against other state-of-the-art appearance-based T\&R systems in outdoor and indoor spaces as well as on the new dataset for appearance-based methods testing presented in \cite{SSM-Nav2}.

This type of T\&R system has many practical applications. Robustness in monotonous environments without an additional infrastructure of visual markers makes them ideal for industrial settings (e.g., autonomous warehouses). Additionally, the option to share the visual information collected in the teaching phase implies simple use in multi-robot systems following the same path.

The rest of the article is structured as follows. Sect. \ref{sec:RW} introduces the related T\&R systems. The new T\&R system and its individual components are described in Sect. \ref{sec:NS}. The results of an experimental evaluation on the dataset and real robot are presented in Sect. \ref{sec:E}. Finally, the achieved results are summarized in Sect. \ref{sec:C}.


\section{RELATED WORK}
\label{sec:RW}
\subsection{Teach-and-Repeat Systems Overview}

As described in the previous section, T\&R systems are classified into position-based and appearance-based. This subsection summarizes the fundamental works from both categories.

A position-based system using SURF \cite{SURF} local visual features capable of following multi-kilometer-long trajectories was presented in 2010 \cite{Barfoot2010}. In following years, various modifications and extensions of this system were published as the system with multi-experience teaching of the trajectory \cite{Barfoot2016_appearance} or the system using deep-learned local visual features \cite{TaR-features}. Contrarily to the appearance-based solutions, all these systems require a stereo camera for creating local 3D maps.

The first appearance-based systems combining robot position tracking along the taught trajectory and reactive control based on stored images emerged in the 1990s \cite{VSRR}. This approach was further improved with the arrival of robust local visual feature detectors. Such a system for polygonal trajectories using SURF features \cite{SURF} was introduced in \cite{Krajnik2010}. Later, it was followed by the BearNav system \cite{BearNav}, which is capable of navigating along general trajectories. It mostly relies on odometry, using the SURF feature matching only to correct drifts and accumulated errors.

An alternative robust shift computation technique not using local visual features is presented in \cite{Rozsypalek-22}. This deep-learning approach employs a Siamese network to encode the information from the compared images into 3D vectors. The best horizontal shift is computed by the cross-correlation of these vectors. This technique serves in new systems, following on BearNav, to directly compute horizontal shifts \cite{Rozsypalek-23} or to support local feature matching \cite{Roucek-22}.

\subsection{Teach-and-Repeat Using Visual Place Recognition}

The main disadvantage of T\&R systems using tracking to localize a robot along the trajectory is that they require starting from the initial position. In contrast, systems performing basic localization can start in the middle of the trajectory and recover from various failures. That motivated the development of T\&R systems using robust VPR methods for basic localization along the trajectory. VPR systems using deep-learning techniques are able to work in changing environments, including day-night or seasonal changes, which is a crucial property for long-term autonomy.

SSM-Nav \cite{SSM-Nav} was the first T\&R system based on an advanced VPR. It uses the Semantic and Spatial Matching Visual Place Recognition (SSM-VPR) \cite{SSM-VPR}. This system performs recognition in two stages, using local visual features extracted from the inner layers of a convolutional neural network. The actual position is estimated by a particle filter that fuses VPR-based localization with wheel odometry. Horizontal shift for steering corrections is computed using shifts between matched local visual features from the second stage of VPR. The system proved the ability to work outdoors and indoors, including challenging lighting conditions or environments with many monolithic surfaces.

A similar experimental system was also developed to navigate a small Unmanned Aerial Vehicle (UAV) \cite{SSM-drone}. As the UAV does not provide reliable odometry information, this system is purely vision-based. Despite that, it proved the ability to navigate in challenging indoor environments.

In 2025, the new version of the SSM-Nav system was introduced \cite{SSM-Nav2}. The main improvements are the multi-platform design, supporting both ground and aerial vehicles, and the employment of new re-ranking techniques (second stage of VPR) introduced in \cite{SSM-VPR2}. The re-ranking is based on D2-Net local visual features \cite{D2-Net}, also used for the horizontal shift computation. The method computes a 2D histogram of shifts from matched features, in which each match votes for its position by a Gaussian function. The main disadvantage of all these systems based on advanced two-stage VPR is the high computational requirement for real-time performance.

\subsection{MixVPR}
\label{sec:RW-MixVPR}

MixVPR \cite{MixVPR} is a modern and efficient system for VPR based on deep learning, achieving performance close to the current state-of-the-art systems \cite{SSM-VPR2}. Contrarily to most of these systems, it performs recognition in one stage instead of two standard steps - filtering and re-ranking. Its main advantages are efficient image representation suitable for large databases and fast recognition.

The system is built on a new neural network architecture combining a standard convolutional backbone with a new function block called Feature Mixer. The entire network computes a global image descriptor,  which is then used to search for the most similar images based on the distances of descriptors. At the beginning of the network, the image is transformed by cropped ResNet network \cite{ResNet} into a~3D tensor. This tensor is sliced into multiple 2D feature maps that are subsequently flattened into 1D vectors. Then, the Feature Mixer transforms these vectors separately by a~cascade of multi-layer perceptron blocks, preserving their dimension. Output vectors are connected into one feature map, whose size is reduced by two fully connected 1D layers applied in horizontal and vertical directions. Finally, the 2D map is flattened and normalized to obtain the image descriptor.

\section{NAVIGATION SYSTEM}
\label{sec:NS}

\subsection{System Architecture}
The presented T\&R system, termed MixTaR, builds on a modified architecture of the recently published SSM-Nav2 system \cite{SSM-Nav2}. The major changes include replacing the original two-stage VPR system with MixVPR \cite{MixVPR} and introducing a new method for horizontal shift computation that uses local features extracted from the inner layers of the MixVPR network. The T\&R system is primarily designed for a~ground mobile robot equipped with a single forward-looking monocular camera.

\begin{figure*}[ht]
  \centering
  \includegraphics[width=\textwidth]{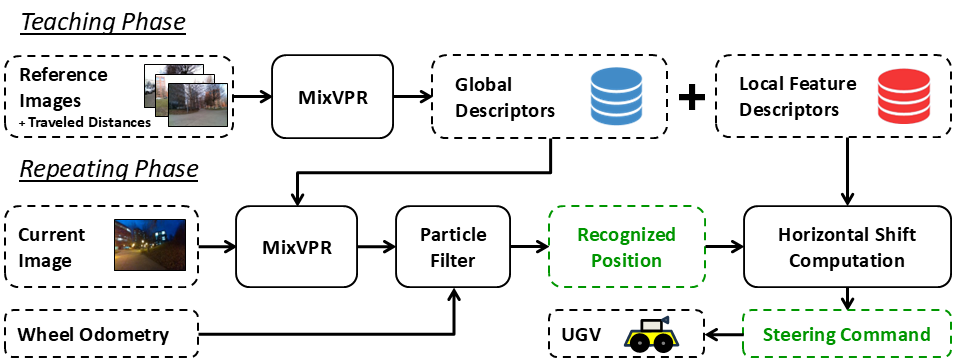}
  \caption{Flowchart of the new T\&R system MixTaR.}
  \label{fig:scheme}
\end{figure*}

The simplified flowchart of the MixTaR T\&R system is presented in Fig.\ref{fig:scheme}, depicting the main operations and results for teaching and repeating phases. During both phases, the system inputs are camera images and wheel odometry, reduced to the traveled distance only. The MixVPR network processes the images to obtain global and local feature descriptors. During the repeating phase, the VPR system uses global descriptors to localize the robot along the taught trajectory. However, the final position is computed by a~particle filter, fusing visual localization with the traveled distance. Local feature descriptors are used to determine the horizontal shift correction, which is then transformed into a~reactive control command for a ground mobile robot. The following subsections describe the individual steps in more detail.

\subsection{Teaching Phase}
During the teaching phase, the robot is navigated by an operator along the desired trajectory while the system autonomously captures and processes images. The system employs the same capturing strategy as \cite{SSM-Nav, SSM-Nav2}, based on the traveled distance measured by odometry. The sampling distance is not fixed but is adjusted according to the current angular velocity. The sampling is denser during turns when differences between consecutive images are significant, and contrarily, it is sparser on straight segments where image differences are minimal. This policy reduces the number of database images and improves localization speed.

Each recorded image is resized to 320 x 320 px resolution  and processed by MixVPR network \cite{MixVPR}, described in Sect.~\ref{sec:RW-MixVPR}. This network was trained by its authors on the GSV-Cities dataset \cite{GSV-Cities} and returns a global descriptor with dimension 4096. An alternative approach to gain local visual features was employed as the spatial information necessary for the horizontal shift computation can not be reconstructed from global descriptors. This method is inspired by older techniques that extract local visual features directly from output tensors of convolutional network inner layers \cite{SSM-VPR, D2-Net}. For that purpose, the new approach uses the MixVPR backbone, which is a cropped ResNet convolutional neural network \cite{ResNet}. The shape of its output is 20x20x1024, and the local features are extracted directly as the normalized columns of this tensor. All local features, the global descriptor, and the traveled distance are stored in the database for each image.

\subsection{Localization and Position Tracking}
Visual localization along the trajectory is ensured by the MixVPR system, searching for the most similar images in a database created during the teaching phase. The best candidates are gained directly based on distances of global descriptors. Equally to the original systems \cite{SSM-Nav, SSM-Nav2}, this position is not used directly to compute the horizontal shift, but it is further processed by a particle filter to suppress incorrectly recognized positions.

This technique was directly adopted from \cite{SSM-Nav2}, so we provide only a brief description here, referring to the original publications for more details \cite{SSM-Nav, SSM-Nav2}. The particles represent 1D positions along the trajectory determined by traveled distances. Visual localization is used to increase the weight of particles close to the recognized candidates, while the traveled distance measured by odometry serves to update the positions of particles as the robot itself moves. The final position is retrieved as the average position of the five best particles. In addition, the particles' weights are used to estimate localization uncertainty.

\subsection{Horizontal Shift Computation}

The horizontal shift is computed between the current camera image and the image closest to the position obtained from the particle filter. The method proceeds from the approach used in the original SSM-Nav system \cite{SSM-Nav} that also uses local features extracted at fixed grid positions.

\begin{figure}[ht]
  \centering
  \includegraphics[width=\linewidth]{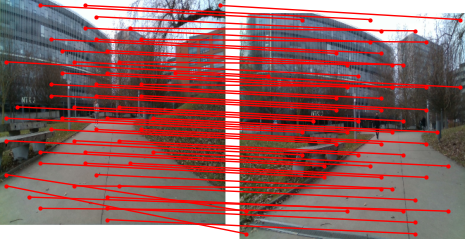}\\
  \vspace{2mm}
  \includegraphics[width=\linewidth]{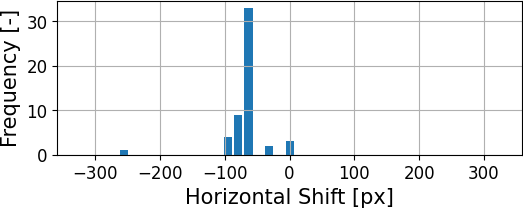}
  \caption{Mutual matches of visual features extracted from the MixVPR backbone (top) and histogram of shifts based on the differences in the matched feature pair coordinates. This visualization contains only a fourth of all real matches.}
  \label{fig:hist}
\end{figure}

At first, the features are matched between images using the standard mutually nearest neighbors approach (cross-matching). This technique substantially filters out wrong matches. Then, the system computes differences in horizontal coordinates (i.e., the shifts) for all found matches and aggregates them into a histogram of all possible shifts. Finally, the shift with the most occurrences is returned as the final one. The approach for one image pair is depicted in Fig.\ref{fig:hist}, including found matches and the histogram.

\section{EXPERIMENTS}
\label{sec:E}

\subsection{Offline Testing}
The new system with MixVPR was first tested using the dataset presented in \cite{SSM-Nav2} and compared with scores of other state-of-the-art systems. They are the SSM-Nav2 system based on D2-Net local visual features \cite{D2-Net}, its earlier version SSM-Nav \cite{SSM-Nav}, and two variants of the BearNav2 system \cite{BearNav2}, using the Siamese network and SIFT local features. The dataset consists of three sequences (outdoors in daylight, outdoors at night, and indoors) and serves for the offline evaluation of the appearance-based systems. The used metrics are the success rate of correct shift estimation and standard deviation. A shift estimated by the T\&R system is considered correct if it lies in the 'ground truth' interval, determined  by the shifts corresponding to the projections of two 3D points located at distances of 5 meters and infinity from the camera origin.

Results displayed in Tab. \ref{Tab:dataset-results} show that standard deviations of the new MixTaR system are comparable with other systems. In terms of the correctness of the shift estimation, MixTaR even outperforms all previous systems.

\begin{table}[b]
    \centering
    \vspace{2mm}
    \caption{Shift computation testing on day-time, night, and indoor sequences of dataset for teach-and-repeat systems \cite{SSM-Nav2}.}
    \renewcommand{\arraystretch}{2.2}
    \begin{tabular}{|c|c|c|c|c|c|c|}
    \hline
     & \makecell{ Correct \\ Shift \\ Day \\{[\%]}} & \makecell{ STD \\ Day \\{[px]}} & \makecell{ Correct \\ Shift \\ Night \\{[\%]}} &  \makecell{ STD \\ Night \\{[px]}} & \makecell{ Correct \\ Shift \\ Indoor \\{[\%]}} &  \makecell{ STD \\ Indoor \\{[px]}} \\ \hline
    \makecell{BearNav2 \\ SIFT} 
    &  99 & 7 & 33 & 233 & 81 & 64 \\ \hline
    \makecell{BearNav2 \\ Siam}
    &  99 & \textbf{6} & 90 & \textbf{17} & 79 &  33\\ \hline
    SSM-Nav &  88 & 9 & 74 & 80 & 68 & 37 \\ \hline
    \makecell{SSM-Nav2 \\ D2-Net}
    & \textbf{100} & 10 & 96 & 19 & 83 & \textbf{27} \\ \hline \hline
    \textbf{MixTaR}
    & \textbf{100} & 8 & \textbf{99} & 22 & \textbf{85} & 36 \\ \hline
    \end{tabular}
  \label{Tab:dataset-results}
  \vspace{-4mm}
\end{table}

\subsection{Hardware and System Setup} 
\begin{figure}[ht]
  \centering
  \includegraphics[width=\linewidth]{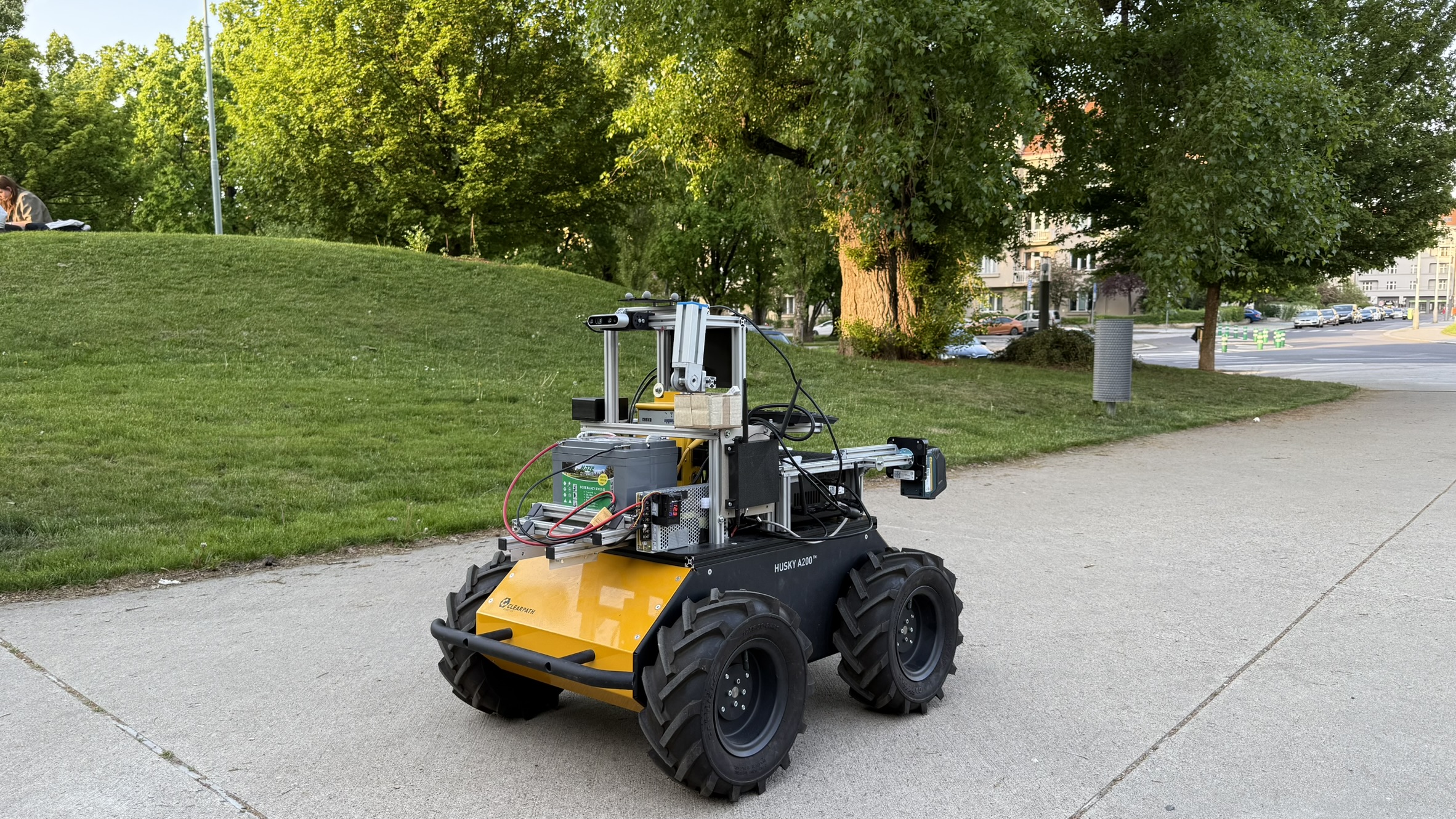}
  \caption{Clearpath Husky A200 robotic platform with additional computational hardware and visual sensor during an outdoor experiment.}
  \label{fig:husky}
\end{figure}

All T\&R systems were tested under real-world conditions on a four-wheeled differential drive robotic platform Clearpath Husky A200, shown in Fig. \ref{fig:husky}. An Intel RealSense D435 camera, offering a maximum resolution of 1920 x 1080 px, was mounted at the front of the robot frame and configured in single RGB monocular mode. An additional external computer was equipped with an Nvidia GeForce GTX 1070 graphics card, an Intel Core i5-7300HQ processor, 32 GB of RAM, and a 500 GB SSD. This setting was identical to the experiments presented in \cite{SSM-Nav, SSM-Nav2}.

The software part of the presented T\&R system was built on ROS2 Humble and Ubuntu 20.04. Individual ROS nodes were developed in Python. The whole platform was adapted for the experiments to run multiple models in parallel during the teaching phase. The user can then decide which system will be used for the navigation in the repeat phase.

\subsection{Indoor Experiments}
Indoor experiments were conducted under controlled conditions in a robotic laboratory equipped with a high-precision Vicon localization system, which was used to measure navigation accuracy. The novel MixTaR system was directly compared to the newest SSM-Nav2 system \cite{SSM-Nav2} and to the more relevant version of BearNav2 \cite{BearNav2} system using the Siamese network. The performance during the repeat phase, in terms of mean and maximum deviation from the taught trajectory, was evaluated on two types of trajectories. The basic trajectory with only 3 turns was 18 meters long, while the more advanced trajectory measured 28 meters. Both trajectories were taught by a human operator for all the systems simultaneously. 

\begin{figure}[ht]
  \centering
  \includegraphics[width=\linewidth]{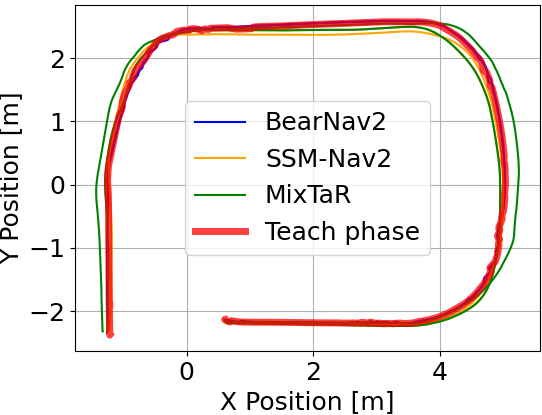}\\
  \vspace{5mm}
  \includegraphics[width=\linewidth]{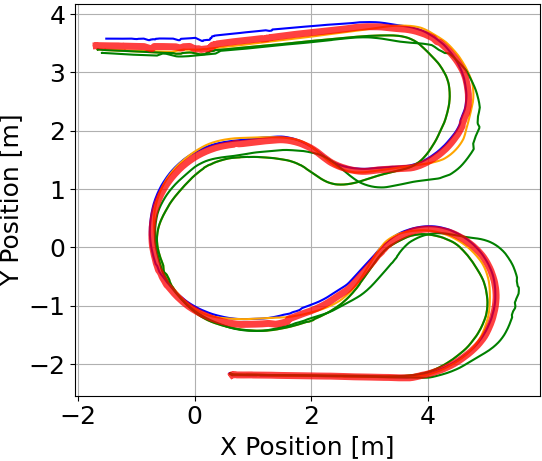}
  \caption{Repeated paths of all tested systems on the basic (top) and advanced (bottom) trajectories.}
  \label{fig:trajectories}
\end{figure}

At first, all systems were tested for navigation from the original starting position under standard lighting. Tab. \ref{Tab:indoor} presents results averaged over two passes for each system. Visual representation of all trajectories recorded by the Vicon system is shown in Fig. \ref{fig:trajectories}. MixTar achieved slightly worse navigation accuracy than both reference systems, but it is still acceptable for many potential applications.

\begin{figure}[ht]
  \centering
  \includegraphics[width=\linewidth]{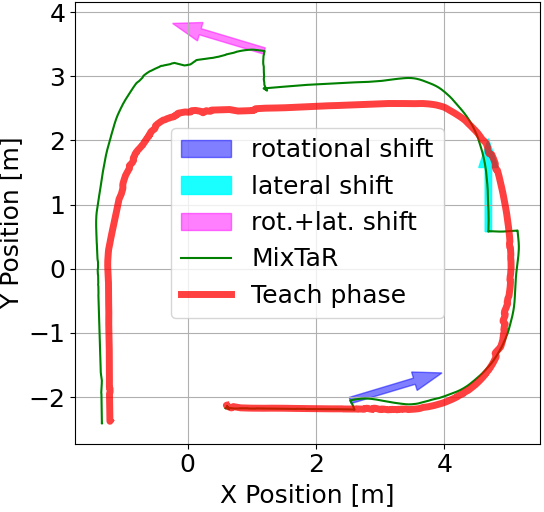}
  \caption{Recorded traversal of the MixTaR system during the experiment with sudden lateral and rotational displacements.}
  \label{fig:kidnapping}
\end{figure}

The second experiment proved the ability of all systems to follow a pretrained trajectory when the starting position was laterally shifted 30 cm to the left. The mean deviations from the taught path are recorded in Tab. \ref{Tab:indoor}. The new MixTaR system was further tested for recovery from sudden lateral and rotational displacements while repeating the basic trajectory, simulating the kidnapping scenario. The resulting path is plotted in Fig. \ref{fig:kidnapping}.

The third part of the experiments was conducted under limited lighting conditions, where only light from outside corridors illuminated the laboratory. All T\&R systems managed to traverse both the basic and advanced path and even maintain performance in terms of mean deviation.

The higher precision of the BearNav system is primarily ensured by repeating the taught velocities. On the other hand, it is unable to localize itself when starting from an arbitrary point along the trajectory. This ability was experimentally verified for both the SSM-Nav2 system as well as for the new MixTaR system.

Moreover, the MixTaR system performs all tasks necessary for appearance-based T\&R navigation in a significantly shorter time than the other system SSM-Nav2. Comparison for both of the systems, in terms of time consumption for feature extraction, VPR candidate detection, and shift computation, is in Tab. \ref{tab:model_times}. Presented times were measured during the repeating phase while driving on the advanced trajectory with more than a hundred images.

\begin{table}[ht]
\centering
\caption{Computational times of SSM-Nav2 and MixTaR systems during navigation on the advanced indoor path.}
\begin{tabular}{|l|c|c|c|}
\hline
Model & \makecell{Feature \\ Extraction (ms)} & \makecell{VPR Candidate \\ Detection (ms)} & \makecell{Shift \\ Computation (ms)} \\ \hline
SSM-Nav2 & 76.5 & 199.8 & 6.7 \\ \hline
MixTaR & \textbf{24.1} & \textbf{36.3} & \textbf{4.9} \\ \hline
\end{tabular}
\label{tab:model_times}
\end{table}

\subsection{Outdoor Experiments}
As a proof of concept, the MixTaR and the SSM-Nav2 systems were tested outdoors on the university campus. Both systems successfully localized themselves along the 150 m long trajectory consisting of hundreds of images and navigated reliably, even with people walking across the camera view. SMM-Nav2 performed with a mean deviation of 11 cm and MixTaR with a mean deviation of \textbf{5} cm, measured at seven points along the trajectory.

\begin{table*}[t]
    \centering
    \vspace{2mm}
    \caption{Navigation precision of T\&R systems for indoor experiments on two testing trajectories, including various lighting conditions and laterally shifted starting position (30 cm).
    }
    \vspace{-2mm}
    \renewcommand{\arraystretch}{1.3}
    \begin{tabular}{|l|c|c|c|c|c|c|c|}
    \hline
    & \multicolumn{4}{|c|}{Basic Path} & \multicolumn{3}{|c|}{Advanced path} \\ \hline
    & \multicolumn{2}{|c|}{Normal Start} & Shifted Start & Night Run & \multicolumn{2}{|c|}{Normal Run} & Night Run \\ \hline
    & \makecell{ Mean \\ Deviation \\{[cm]}} & \makecell{ Maximum \\ Deviation \\{[cm]}} & \makecell{Mean \\ Deviation \\ {[cm]}} & \makecell{Mean \\ Deviation \\ {[cm]}} &  \makecell{ Mean \\ Deviation \\{[cm]}} & \makecell{ Maximum \\ Deviation \\{[cm]}} & \makecell{ Mean \\ Deviation \\{[cm]}} \\ 
    \hline
    BearNav2 Siam & \textbf{1.4} & \textbf{6.5} & 13.4 & \textbf{1.5} & \textbf{5.0} & \textbf{17.4} & \textbf{6.9} \\
    \hline   
    SSM-Nav2 (D2-Net) & 5.3 & 16.8 & \textbf{9.6} & 9.5 & 8.4 & 24.0 & 12.2 \\ \hline
    MixTaR & 15.4 & 40.0 & 12.4 & 18.0 & 12.2 & 38 & 21.1 \\ \hline
    \end{tabular}
  \vspace{-5mm}
  \label{Tab:indoor}
\end{table*}

\section{CONCLUSIONS}
\label{sec:C}
The new MixTaR system shows qualities necessary for long-term autonomy in mobile robotics. It reliably navigates in challenging environments with moving people or with variable lighting conditions. Using a modern and efficient MixVPR system for localization, it runs faster than other appearance-based T\&R systems of this type.

On a novel dataset for testing horizontal shift computation methods used in appearance-based T\&R systems, it outperformed other state-of-the-art approaches in terms of the percentage of successfully calculated shifts. Real-world experiments proved that the system follows taught paths with the accuracy comparable to other systems. Moreover, it is capable of localizing itself from an arbitrary point on the trajectory and completing the task.

The speed and efficiency of the presented system make robust and quality T\&R navigation suitable for smaller mobile robots with limited hardware capabilities, such as UAVs. Deployment on low-end hardware and subsequent extensive experiments are the next steps in the future development of the MixTaR system.








\bibliographystyle{IEEEtran}
\bibliography{bibliography}

\end{document}